\documentclass[twocolumn]{ceurart}
\usepackage{listings}
\usepackage{hyperref}
\usepackage{comment}
\usepackage{graphicx}
\usepackage{subcaption}
\begin{document}

\conference{Workshop on Artificial Intelligence and Robotics (AIRO 2026 @ AIxIA 2026)}

\title{Skill Composition for Legged Robot Reinforcement Learning}

\author[1]{Daniel Gigliotti}[%
  orcid=0009-0002-1485-5539,
]
\fnmark[1]

\author[1]{Flavio Maiorana}[%
  orcid=0009-0003-2059-7254,
  email=maiorana@diag.uniroma1.it,
]
\fnmark[1]

\author[1]{Fabio Patrizi}[%
  orcid=0000-0002-9116-251X,
  email=patrizi@diag.uniroma1.it,
]

\author[1]{Luca Iocchi}[%
  orcid=0000-0001-9057-8946,
  email=iocchi@diag.uniroma1.it,
]

\address[1]{
  Dept. of Computer, Control, and Management Engineering\\ Sapienza
  University of Rome, Rome (Italy)
}

\cortext[1]{Corresponding author.}
\fntext[1]{These authors contributed equally.}

\begin{abstract}
Robots, and humanoid robots in particular, are increasingly competent at individual behaviors, each obtained by training a specialized controller. A specialized skill is quick to train, converges reliably because the problem it faces is narrow, and can be validated on its own, none of which is true of a single end-to-end policy asked to cover everything. What remains fragile is the transition between them. We argue that the composition of independent sub-policies deserves to be treated as a research problem in its own right, rather than as an implementation detail left to whatever mechanism happens to be at hand.


Reliable composition is what turns a collection of separate skills into a repertoire that can be used, extended and shared. More fundamentally, if control can be passed between specialized policies safely, and at any moment, the choice of what the robot should do next can be delegated to a component of an entirely different nature, such as a planner, an automaton or a symbolic controller, whose behavior can be inspected in advance. The policies would then only ever have to act, and what the robot can be trusted to do would become verifiable.
\end{abstract}


\maketitle

\section{Introduction}
\label{sec:intro}

Learning individual sensorimotor skills on legged and humanoid robots has become remarkably effective. Deep Reinforcement Learning (DRL) in simulation, followed by transfer to hardware, now produces agile behaviors that were out of reach a few years ago: parkour over rough terrain \citep{zhuang2024humanoid}, blind locomotion in natural environments \citep{lee2020quadruped}, dynamic striking in robot soccer \citep{xu2026humanoidsoccer}.

These results demonstrate that a single policy, optimized against a task reward, can uncover sophisticated behavior. Each of them, though, is one skill, trained and validated inside one regime. How to obtain a robot that handles a genuinely complex task, spanning several such skills, remains open. Several approaches have been proposed, each with its own limitation.

\textbf{Model-based control} derives whole-body optimal or predictive controllers from a model of the robot's mechanics. It requires no training data and admits formal guarantees. Every new behavior, however, calls for new modeling and new tuning, and the contact-rich interactions of unstructured environments remain hard to model.

\textbf{End-to-end learning} imposes no decomposition at all, and is the most common approach today: a single policy is optimized against a global task reward, and is expected to acquire on its own every competence the task requires. Three difficulties follow. The reward must encode stability, task progress and safety at once, and balancing these terms against one another is a problem in itself. Credit assignment degrades over long horizons, where the actions responsible for an outcome are far removed from it. And nothing is reusable: since every competence resides in the same set of weights, acquiring a new one means training the policy again from scratch. Such a policy is also opaque. Every competence is entangled in the same network, so no part of it can be studied in isolation.

\textbf{Modular skills} offer a compelling alternative, in which the competences are kept separate. Each behavior is learned as its own policy, trained on its own regime, and a complex task is carried out by combining several of them. A policy of this size is inexpensive to train, converges reliably because the problem it faces is narrow, and can be validated in isolation. None of this holds for a single policy asked to cover everything.

The decomposition is most valuable when the experts remain unchanged once trained. A frozen expert is developed and validated in isolation, and then added to the repertoire without revisiting the skills already in it, so that the repertoire grows by addition. We therefore treat the preservation of the experts as a design objective, to be relaxed only where necessary. Much of the existing literature on composition instead finetunes the skills so that they fit one another, recovering robustness at the cost of the modularity that motivated the decomposition in the first place.

What the decomposition leaves open is how the experts are combined, and we identify two mechanisms for doing so. Both rely on a decision layer, separate from the experts, that determines how much each of them contributes; what differs is what that layer produces. In \textbf{blending}, it assigns a weight to each expert, and their outputs are combined into a single action; the weights vary continuously, so a change of regime is absorbed by the merge as it happens. In \textbf{bridging}, it names a single expert, which remains in control until another is named; the change is abrupt, and something must cover the moment at which it occurs, since the outgoing expert may leave the robot in a state from which the incoming one cannot succeed. Neither mechanism subsumes the other, but the two decisions are of different kinds. A weighting over experts is a continuous quantity, tied to the robot's dynamics; a choice among named skills is not, and can therefore be delegated to a component selected for its interpretability, such as a planner, an automaton or a symbolic controller, whose decisions can be accounted for independently of the policies that carry them out, rather than a black box whose choices can be examined only after the fact.

Bridging, in itself, is not a new idea. In the options framework \citep{sutton1999options} each skill declares the set of states from which it may be initiated and a condition under which it terminates; composing two skills then reduces to arranging for the first to terminate inside the initiation set of the second. Our setting removes this guarantee. The decision to switch does not belong to the skill that is running and does not wait for it: it originates outside, at an arbitrary moment. The state in which control changes hands is therefore not one the outgoing skill was ever required to reach, and nothing ensures that it belongs to the initiation set of the incoming one.

One remedy is always available, and is what most systems do: bring the robot to rest, and invoke the next skill from a state its initiation set is guaranteed to contain. This reduces a repertoire to a collection of behaviors that can only be executed one at a time, since the momentum accumulated by the outgoing skill is discarded when it could have been exploited instead. A humanoid required to jump while running is a clean example. The jump is feasible from a standstill, so a robot that stops, jumps and resumes composes the two skills correctly. What it gives up is the momentum already built up, which a jump taken on the run would have turned to its advantage.

Our position is that composition deserves to be treated as a research problem in its own right, rather than as an implementation detail settled by whatever mechanism happens to be at hand. We state the composition problem (Section~\ref{sec:problem}), argue what a good solution would enable (Section~\ref{sec:enable}), review the literature (Section~\ref{sec:related}), identify what we believe is open (Section~\ref{sec:challenges}) and report first results on both mechanisms (Section~\ref{sec:results}).

\section{Problem definition}
\label{sec:problem}

We define our problem as a contextual Markov Decision Process (CMDP) \citep{hallak2015contextual} $\mathcal{M} = \langle \mathcal{S}, \mathcal{A}, \mathcal{X}, P, R, \gamma \rangle,$ in which a context $x \in \mathcal{X}$ parametrizes the task reward $R(s,a,x)$. The context stands for any parametrization of the task, such as a velocity command, a target position or a higher-level instruction, with goal reaching being the special case in which $x$ is a target state. We are interested in tasks whose context decomposes into $N$ regimes that can be pursued independently, $x = (x_1, \dots, x_N) \in \mathcal{X}_1 \times \cdots \times \mathcal{X}_N,$ with reward $R(s,a,x) = \hat{C}(R_1(s,a,x_1), \dots, R_N(s,a,x_N))$ for some combination rule $\hat{C}$. A \emph{skill} (or \emph{expert}) is a sub-policy $\pi_i(a \mid s, x_i)$, one per regime and indexed by $i \in \{1,\dots,N\}$, optimal for its own regime reward $R_i$ and indifferent to the others. Variation within a regime, such as a change in the commanded velocity, is handled by the skill's own generalization; variation across regimes changes which reward term, and correspondingly which skill, governs behavior.

Given a complex task, we seek $N$ skills $\{\pi_i\}_{i=1}^N$, a decision layer that determines their involvement over time, and a composition operator $C : \Pi_1 \times \cdots \times \Pi_N \to \Pi,$ mirroring $\hat{C}$ at the policy level, such that $\pi = C(\pi_1, \dots, \pi_N)$ solves the original task. The central question this raises, and which motivates the rest of this paper, is how $\{\pi_i\}_{i=1}^N$ and $C$ should be learned so that $\pi$ recovers, or exceeds, the performance of the end-to-end solution while retaining reusability of $\{\pi_i\}_{i=1}^N$ across tasks.

\subsection{Two ways of composing experts}
\label{sec:composing}

We identify two ways to instantiate the composition operator $C$ around the decision layer, corresponding to the two forms $\hat{C}$ can take. They differ in what the decision layer emits, and in what is required of the composition when its output changes.

The first is \textbf{blending}, instantiating $C$ as a \emph{merge} $f$ of concurrently active experts, weighted by a decision layer $\omega : \mathcal{S} \times \mathcal{G} \to \Delta^{N-1}$ that assigns each expert a share of the command:
\[
C_f(\pi_1,\dots,\pi_N)(a \mid s,g) = f\big(\omega(s,g);\, \pi_1(a\mid s,g), \dots, \pi_N(a\mid s,g)\big).
\]
This is attractive for tasks that are genuinely the superposition of two competences, loco-manipulation being the obvious case, where one expert could handle the legs and another the arms. Formally, nothing guarantees that $f$ is competent in states where both of its inputs are competent in isolation; in practice, $f$ can itself be learned, familiarizing the merged command with training data from the simulator. Because $\omega$ varies continuously, a change of regime is absorbed by the merge itself, and no further mechanism is required at the moment the involvement of the experts changes.

The second is \textbf{bridging}, instantiating $C$ as a decision layer $\sigma : \mathcal{S} \times \mathcal{G} \to \{1,\dots,N\}$ that selects a single active expert:
\[
C_\sigma(\pi_1,\dots,\pi_N)(a \mid s,g) = \pi_{\sigma(s,g)}(a \mid s,g).
\]
Here the change is abrupt, and a mechanism is required precisely at the moment it occurs. Between two experts, a short-lived connective behavior is inserted, which we call a \emph{bridge} $\beta$. While an expert is running, $\sigma$ signals that a different expert should take over; the bridge takes control immediately, and its single responsibility is to make the incoming expert able to take over, from whatever state the outgoing one has left the robot in. Once the handover is done, the bridge disappears, so that, unlike $\pi_i$, $\beta$ is not itself a solution to any regime's reward $R_i$, but a transient policy defined only by the handover it enables.

\subsection{What a good composition method would enable}
\label{sec:enable}

Composition would change what can be built out of a set of skills.

Blending would yield behaviors that none of the experts possesses on its own. Loco-manipulation is the standing example: a locomotion expert and a manipulation expert, neither of which can carry out the task alone, together produce a robot that manipulates an object while walking. 

Bridging would extend what can be built in the other direction. Most tasks worth doing are longer than any single skill: they are successions of competences, and every junction between two of them is a point at which the whole task can fail. Were those junctions reliable, a long-horizon task would become largely a matter of specifying a high level sequence of instructions. 

Under either mechanism the library grows by addition, a new behavior being obtained by training one policy and connecting it to those already there.

Reliable bridging would also change who decides what the robot does next. There is considerable current interest in delegating that decision to a large language or vision-language model \citep{ahn2022saycan,kim2024openvla}. Such a module is powerful but opaque: we cannot say in advance what it will choose, and can only partially reconstruct, afterwards, why. The alternative is a decision layer assembled from components whose behavior can be characterized before it runs, such as a planner, an automaton or a neuro-symbolic controller, which derive a decision from stated criteria and can be held to it. What has kept such components at arm's length from learned control is not a lack of expressiveness, but the gap between deciding and acting: a decision issued at an arbitrary moment, by a component that is not required to reason about the robot's dynamics, must still be carried out by a machine in whatever state it happens to be in. A bridge is what spans that gap. The decision it asks for is a choice among named skills rather than a continuous weighting; and since the bridge takes whatever state the robot is in when the decision arrives, that decision need not be correct about the dynamics in order to be safe to act on.

Safety follows from the same arrangement. A component whose explicit and only job is to ensure that the incoming expert can take over, whatever state the outgoing one has left behind, is a natural place to state and check safety conditions, and a far more tractable one than a controller that does everything at once.

\section{Related work}
\label{sec:related}

Bridging two skills has a recognizable literature, grouped here into four research lines. All of it assumes the classical account recalled in Section~\ref{sec:intro}, in which the outgoing skill is allowed to terminate; none addresses the case in which it is interrupted instead, so that nothing is guaranteed about the state the bridge inherits.

The first line learns the bridge. \citet{lee2019transition} introduces transition policies, short policies joining one skill to the next; feedback is sparse, arriving only once the next skill returns a reward, so they add an estimate of how promising each state is as a starting point. \citet{byun2022distribution} instead trains the bridge to be indistinguishable from the next skill's own behavior near its usual starting point, and separates out the decision of when to hand over. Others reshape rewards and replay failures at skill boundaries \citep{xin2026seamless}.

The second line adjusts the skills so that they fit together. \citet{lee2021tstar} finetunes each skill so that where it ends falls inside where the next can start, since widening every skill's acceptable starting conditions does not scale with the length of the chain; \citet{chen2023seqdex} and \citet{chen2024scar} learn whether one skill can follow another and adjust the skills accordingly, while residual learning freezes the expert and trains an additive correction on top of it \citep{silver2018residual,johannink2019residual}. This line has the strongest published results on long horizon manipulation, but buys robustness by relaxing the independence of the experts.

The third line concerns the moment of the handover. Closest to what we want, and on a biped, are the setup policies of \citet{tidd2022setup}, which bridge from default walking to a terrain specific skill and pick their own switching moment, reusing the target skill's training-time assessment of a state, discounted wherever it can be shown unreliable. \citet{chai2025n2m} learn which positions a manipulation policy prefers to be started from and drive the robot there first, and \citet{yu2026skillswitch} discover the criteria for switching between quadruped gaits rather than fixing them by hand.

The fourth line is recent work on humanoids. Two lines blend rather than switch: \citet{kuang2025skillblender} learn a high level policy combining the outputs of frozen primitives, and \citet{xin2026rpg} train each expert under stochastic interruption and blend them through a learned gate. A third is retrieval based: \citet{wu2026parkour} match the robot's state against frames of a motion database, posing the question of where to aim as a matching problem.

Two observations follow. First, no existing method simultaneously keeps the experts frozen, bridges between them, and allows the switch to be requested at an arbitrary moment: the bridging works wait for a natural end or a scripted interval, while the work that does interrupt either widens the experts or blends them. Second, and more importantly, nobody chooses where to land: the third line decides at which moment to stop bridging, never which state of the next skill's initiation set is the best one to aim for.

\section{Open challenges and directions}
\label{sec:challenges}

A bridge that can only choose a moment cannot exploit the momentum the robot built up: it can delay the handover until the state becomes acceptable, but not aim for a different acceptable state that would serve the next skill better. The robot that must jump while running is the clearest case: the jumping skill starts from a whole family of states, some throwing away the motion already built up and others exploiting it. Choosing among them is a question of where, not of when, and one that existing methods do not address.

Two things are missing before such a choice can be made. One is a means of \emph{comparison}: saying that one state is a better place to hand over than another presupposes a way of comparing them, and the obvious measures do not serve, since two states can be close by any ordinary account and still lead to entirely different outcomes. Two states should count as close when the same behavior, applied to both, leads to comparable outcomes, and distances of this kind have been learned, though for other purposes \citep{zhang2021bisim}. The other is \emph{anticipation}: to choose a landing state, the bridge must know what follows from it. This suggests giving each expert a compact descriptor of the behavior it would produce shortly after taking control from a given state. Such a descriptor belongs to the expert rather than to the pair, so it is computed once, at training time, and consulted by any bridge, turning the choice into a comparison between predictions rather than a search by trial.

How a bridge should be organized is open in a different way. Existing methods attach one bridge per skill, or one per pair, which does not obviously scale and forgoes any sharing, since much of what a bridge must know, such as keeping a humanoid upright or shedding momentum, is not specific to the pair it connects. Whether a single bridge, told which expert is to be entered, can match a family of specialized ones is another question worth answering.

There is also the matter of \emph{interruption}. A decision layer that reacts to the world cannot wait for the current skill to reach a natural end, and a skill interrupted in the middle is both where the state handed to the bridge is least predictable and the case that existing work does not cover.

Blending raises a different question. Nothing guarantees that a merge of two experts is competent in the states where each of them is competent alone, and the difficulty grows as their action spaces overlap: two experts commanding disjoint parts of the body interfere little, whereas two that drive the same joints may jointly produce a command neither would have chosen. Characterizing the states in which a blend can be trusted, without retraining the experts it combines, is equally open.

None of this can be settled without \emph{measurement}. Success counts on long-horizon tasks conflate the quality of the transitions with that of the skills and of the decision layer; isolating the transition, by measuring how often a handover succeeds, what it costs in time and motion, and how far the robot had to be displaced, would make results comparable across methods.

\section{First Results}
\label{sec:results}

We are pursuing both of the arrangements of Section~\ref{sec:problem}, one contribution each: a blended composition on a ball-kicking task, and a bridged composition on a set of testbeds.

\subsection{Blending: kicking as a residual on locomotion}
As a concrete instance of skill fusion, we applied a hierarchical approach to a ball-kicking task on a humanoid robot. We instantiate the blend $C_f$ of Section~\ref{sec:composing} with $N=2$ experts: a frozen locomotion policy $\pi_{\text{loc}}(a \mid s, x_{\text{loc}})$, trained on velocity-command tracking, and a kicking policy $\pi_{\text{kick}}(a \mid s, x_{\text{kick}})$ (the context corresponds here to the desired kicking direction), trained as a \emph{residual} on top of it, following a variant of the approach of \citet{kumar2023cascaded}. Each expert is a Gaussian policy $(\mu_i, \sigma_i)$ over continuous joint-space actions, the standard parametrization for PPO in continuous control; a learned \emph{orchestrator} network $\omega$ observes task-relevant state (including the robot's distance to the ball) and outputs softmax blend weights $w = \omega(s) \in \Delta^{N-1}$. $f$ is realized as a weighted blend of the Gaussians \[\sigma = \left(\sum_{i} \frac{w_i}{\sigma_i}\right)^{-1}\!\!, \quad
\mu = \sigma \sum_{i} \frac{w_i \mu_i}{\sigma_i}\] and the blended action is sampled as $a \sim \mathcal{N}(\mu, \sigma)$. Unlike \citeauthor{kumar2023cascaded}'s global weighting, $w$ is predicted independently per joint. The orchestrator is initialized to favor $\pi_{\text{loc}}$ almost exclusively ($w \approx (0.98, 0.02)$), and $\pi_{\text{kick}}$'s action magnitude and blend weight are both penalized during training, so the residual only gains authority where it earns it, based on the rewards valid for the kick's policy regime. Notably, the orchestrator receives the robot's distance to the ball as part of its observation, and we observe empirically that it learns to favor $\pi_{\text{kick}}$ only in proximity to the ball, recovering the intuitive far/near behavior without it being imposed as a rule. Only $\pi_{\text{kick}}$ and the orchestrator are trained; $\pi_{\text{loc}}$ remains frozen throughout. Rewards are split into groups, each assigned its own critic and GAE advantage; the per-group advantages are normalized independently and summed, following the multi-critic actor learning paradigm \citep{mysore2022multicritic}. The robot is additionally trained to recover its equilibrium after kicking, resuming locomotion. This return bridge required no explicit training of its own: $\pi_{\text{loc}}$'s domain randomization is broad enough that the post-kick state distribution already falls within the states it is trained to recover from. Read against Section~\ref{sec:related}'s second observation, this sidesteps the choose-where-to-land problem rather than solving it, by making the target skill's acceptance region wide enough, via randomization, to absorb the bridge's endpoint without an explicit landing objective. We report no quantitative results at this stage; our contribution is the composition mechanism $C_f$ as instantiated above and a working implementation of it, intended as a baseline for later refinement.

To illustrate the effect of the composition mechanism itself, a variant with locomotion and kicking treated as separate behaviors switched between rather than fused is in \href{https://youtube.com/shorts/7BwfZ36myvE?is=5zk04m7ih2eDOEBW}{this video}. Unlike the residual formulation above, the transition here is visibly discontinuous: because the kicking skill is only ever exposed to standing initial states during training, invoking it while the robot is mid-stride lands it outside its training distribution, producing a visible break at the switch.

\begin{figure}[t]
    \centering
    \begin{subfigure}[b]{0.48\linewidth}
        \centering
        \includegraphics[width=\linewidth]{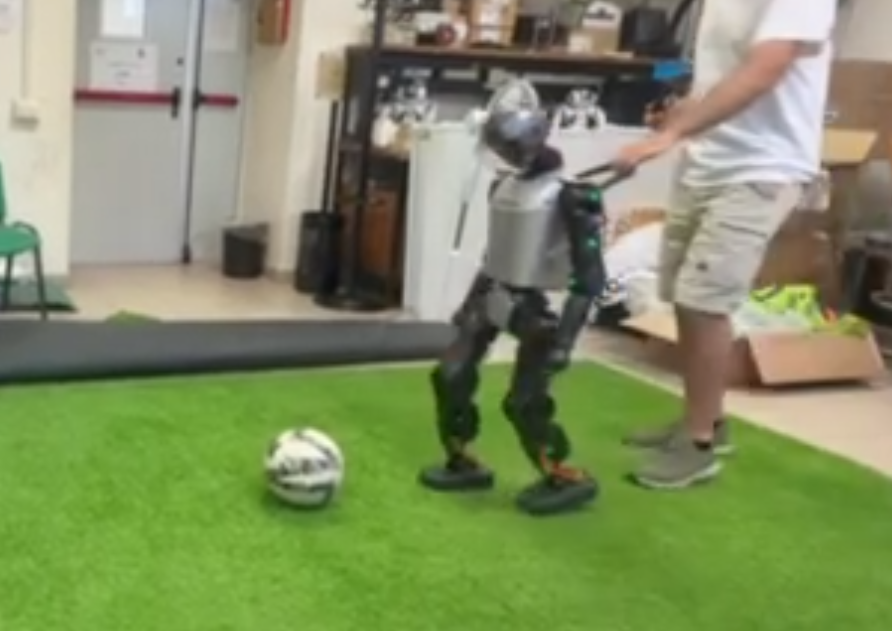}
        \caption{Locomotion regime.}
        \label{fig:loco}
    \end{subfigure}
    \hfill
    \begin{subfigure}[b]{0.48\linewidth}
        \centering
        \includegraphics[width=\linewidth]{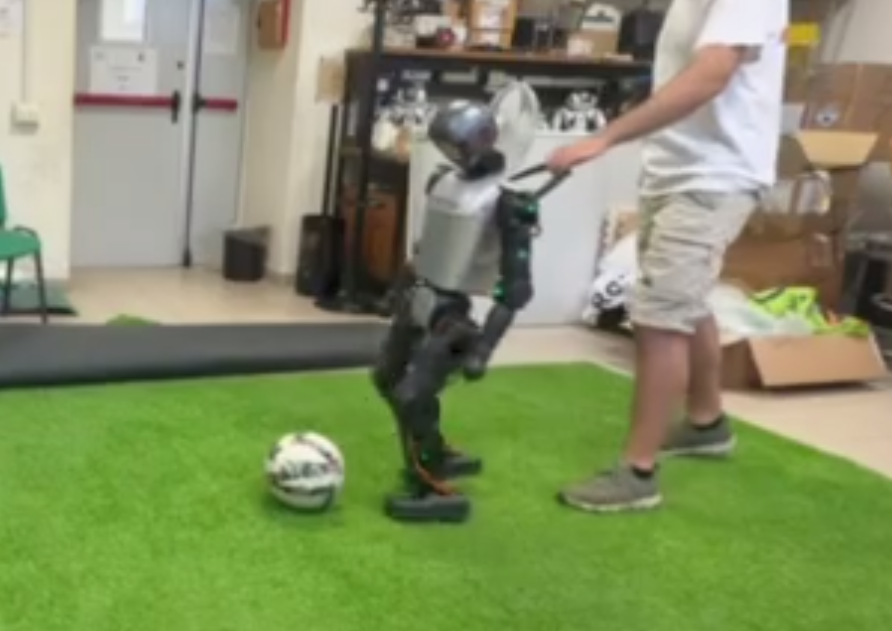}
        \caption{Adjusting for the kick.}
        \label{fig:adjust}
    \end{subfigure}

    \vspace{0.5em}

    \begin{subfigure}[b]{0.48\linewidth}
        \centering
        \includegraphics[width=\linewidth]{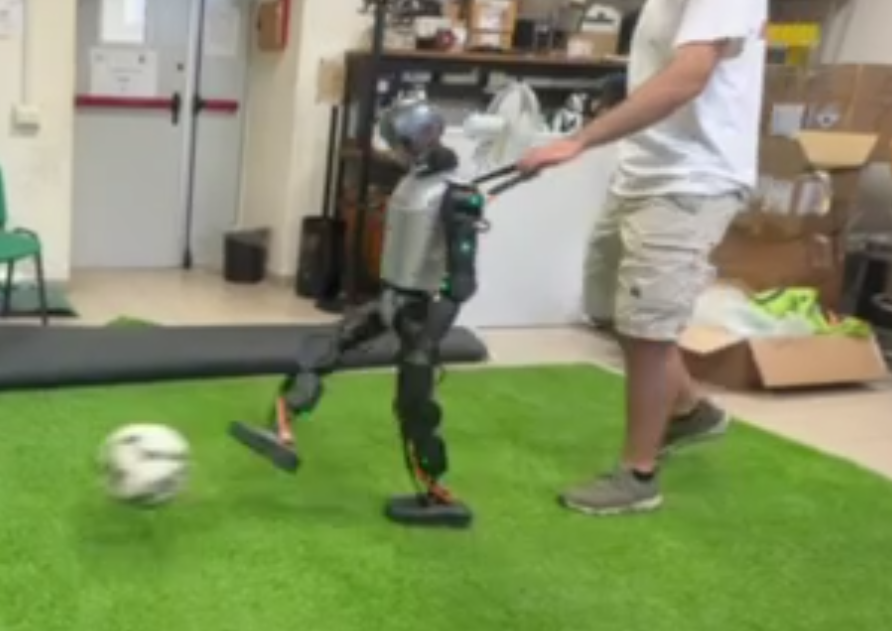}
        \caption{Blended Kick action.}
        \label{fig:kick}
    \end{subfigure}
    \hfill
    \begin{subfigure}[b]{0.48\linewidth}
        \centering
        \includegraphics[width=\linewidth]{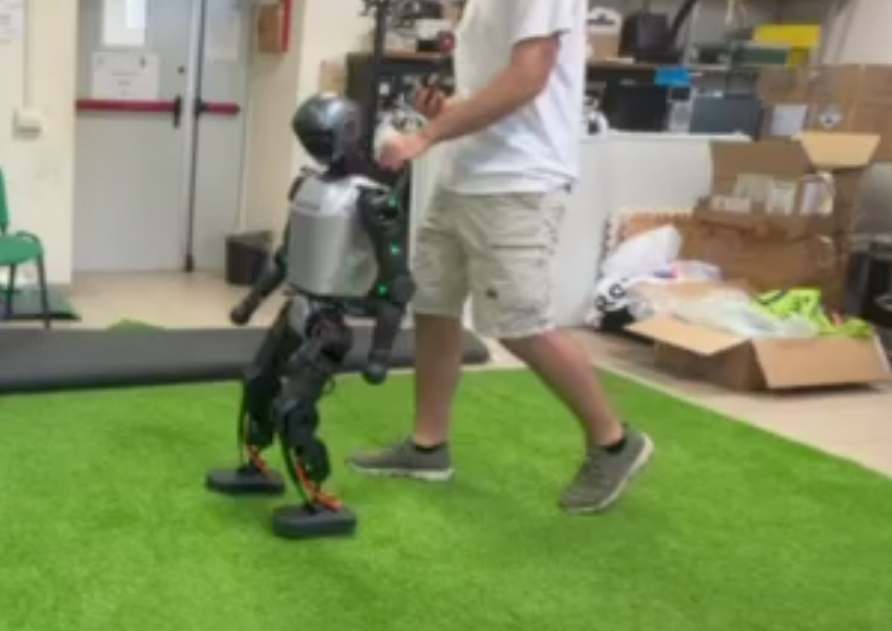}
        \caption{Post-kick recovery.}
        \label{fig:recover}
    \end{subfigure}

    \caption{Overview of the kick divided in four stages. \href{https://drive.google.com/file/d/1Kv2GMaUsXif6x9r6RHa9PGjZd5UhorMD/view}{Complete video}
    }
    \label{fig:overview}
\end{figure}

\subsection{Bridging: jumping while walking, and other testbeds}
\label{sec:bridging}

Our second contribution concerns bridging, and takes up one of the challenges raised in Section~\ref{sec:challenges}: it uses a single bridge for a whole set of skills.

Our main testbed is a humanoid robot running a corridor and alternating between a frozen walking skill and a frozen jumping skill. The jumping skill was trained from rest, so the states it is known to start from form a narrow set around a standstill crouch, and this is what makes the handover hard. Handing the jump a robot that is still moving fails outright: the skill is asked to reproduce a crouch it has no way of reaching from where the robot is. Bringing the robot to a stop first does succeed, but only by dissipating the momentum that made the jump worth attempting in the first place. Our bridge does neither. It carries the robot from walking into a state the jump can in fact be entered from, one that lies outside the narrow set the skill was trained on and that retains the motion already built up.\footnote{\url{https://drive.google.com/file/d/193oTq-TE8TrzKpzx3t5-XUZ5XgterVSA/view?usp=sharing}}

\begin{figure}[t]
    \centering
    \begin{subfigure}[b]{0.48\linewidth}
        \centering
        \includegraphics[width=\linewidth]{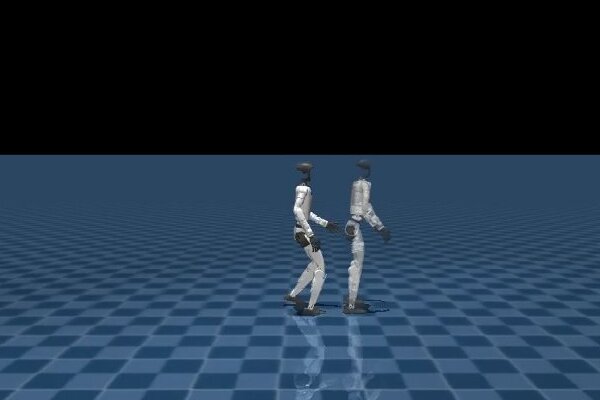}
    \end{subfigure}
    \hfill
    \begin{subfigure}[b]{0.48\linewidth}
        \centering
        \includegraphics[width=\linewidth]{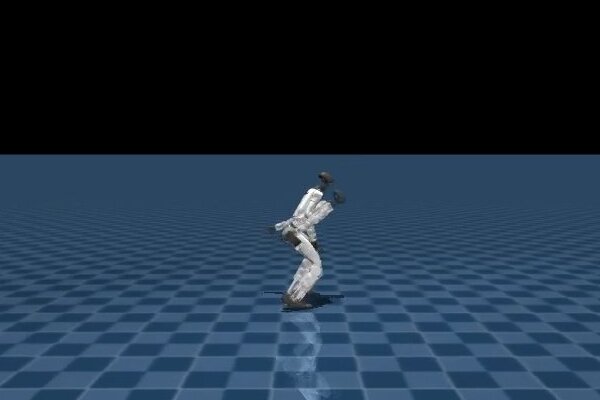}
    \end{subfigure}

    \vspace{0.5em}

    \begin{subfigure}[b]{0.48\linewidth}
        \centering
        \includegraphics[width=\linewidth]{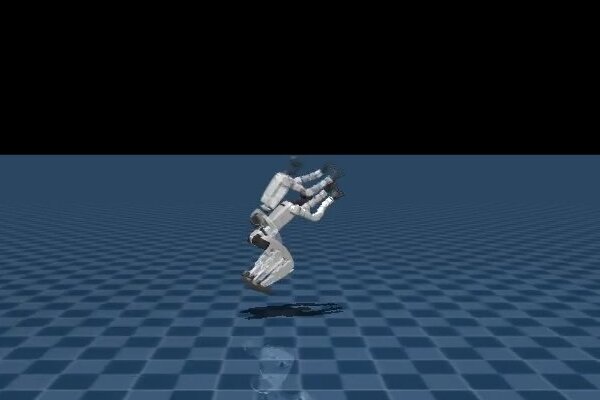}
    \end{subfigure}
    \hfill
    \begin{subfigure}[b]{0.48\linewidth}
        \centering
        \includegraphics[width=\linewidth]{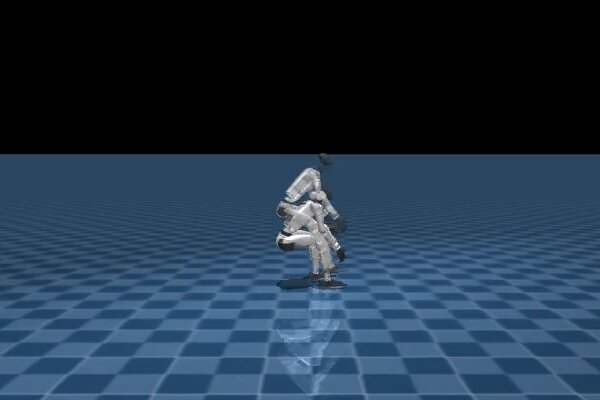}
    \end{subfigure}

    \caption{A humanoid robot (Unitree G1) switching from walking to jumping while exploiting the momentum it has built up. It does not stop walking before jumping: it transitions directly into a crouching state from which the jumping skill can take control. The ghost shows the nominal execution of the jump, as it would be performed if started from the states the skill was trained on.}
    \label{fig:jump}
\end{figure}

We describe the mechanism only as far as is needed here. The bridge is trained once, under physics, on gaps cut out of a corpus of recorded human motion \citep{harvey2020lafan1}: it is placed at the frame where the recording was interrupted, momentum included, told where it must arrive, and never shown what went in between. Some of these gaps are spliced, their two halves taken from different recordings, so that the bridge is asked to join two motions rather than to continue one. It is scored both on the state it hands over in and on what happens over the stretch that follows, which prevents a plausible-looking crossing that arrives in the wrong state from counting as a success. No skill is further trained to fit the others: the experts stay frozen and independent, and the bridge does not have to be retrained when the set of skills changes. Which state to aim at is, for now, chosen from a small set of entry states identified per skill; learning that choice is the immediate next step.

Two smaller testbeds put the same arrangement on robots for which no corpus of recorded motion exists, and were chosen because each isolates a different reason a handover can fail. On a \textbf{cart-pole}, a swing-up skill pumps energy into the pole but cannot balance it, while a balancer holds it upright but is valid only in a narrow band around vertical. The two have to be composed, and the instant at which control passes decides the outcome: too early or too late and the pole is lost. What the testbed shows is that the bridge can identify and reach the state the balancer succeeds from.\footnote{\url{https://drive.google.com/file/d/1v7XovVJEcEKU6oxpJ3has8ebmCluWHw3/view?usp=sharing}}

On a \textbf{differential-drive robot}, a fast straight-line drive is composed with a tight low-speed turn that is safe at the speed it was trained at and rolls the chassis over if it is handed the robot at cruise. Here the difficulty is not timing but speed: the bridge has to shed exactly as much of it as the turn requires.\footnote{\url{https://drive.google.com/file/d/1iBmuqi9um7YX4yfuNmGjeufppxwauuHt/view?usp=sharing}}

Neither robot has a corpus of recorded motion to draw on, so we recorded one: a few seconds of rollout from the skill being entered supplies the motion the bridge is aimed at. Everything else is unchanged.

Results at this stage are qualitative, and the recordings linked above are the report. A quantitative account of the three testbeds, and the choice of where to land as a learned component rather than a fixed one, is what we intend to bring next.

\bibliography{ref}

@String{Chelsea = "Chelsea" }

@inproceedings{zhuang2024humanoid,
  author    = {Zhuang, Ziwen and Yao, Shenzhe and Zhao, Hang},
  title     = {Humanoid Parkour Learning},
  booktitle = {Conference on Robot Learning (CoRL)},
  year      = {2024}
}

@article{lee2020quadruped,
  author  = {Lee, Joonho and Hwangbo, Jemin and Wellhausen, Lorenz and
             Koltun, Vladlen and Hutter, Marco},
  title   = {Learning quadrupedal locomotion over challenging terrain},
  journal = {Science Robotics},
  volume  = {5},
  number  = {47},
  pages   = {eabc5986},
  year    = {2020},
  doi     = {10.1126/scirobotics.abc5986}
}

@inproceedings{xu2026humanoidsoccer,
  author    = {Xu, Zifan and Seo, Myoungkyu and Lee, Dongmyeong and Fu, Hao and
               Hu, Jiaheng and Cui, Jiaxun and Jiang, Yuqian and Wang, Zhihan and
               Brund, Anastasiia and Biswas, Joydeep and Stone, Peter},
  title     = {Learning Agile Striker Skills for Humanoid Soccer Robots from
               Noisy Sensory Input},
  booktitle = {IEEE International Conference on Robotics and Automation (ICRA)},
  year      = {2026},
  note      = {arXiv:2512.06571}
}

@inproceedings{ahn2022saycan,
  author    = {Ahn, Michael and Brohan, Anthony and Brown, Noah and
               Chebotar, Yevgen and Cortes, Omar and David, Byron and
               Finn, Chelsea and Fu, Chuyuan and Gopalakrishnan, Keerthana and
               Hausman, Karol and Herzog, Alex and Ho, Daniel and
               Hsu, Jasmine and Ibarz, Julian and Ichter, Brian and
               Irpan, Alex and Jang, Eric and Ruano, Rosario Jauregui and
               Jeffrey, Kyle and Jesmonth, Sally and Joshi, Nikhil J. and
               Julian, Ryan and Kalashnikov, Dmitry and Kuang, Yuheng and
               Lee, Kuang-Huei and Levine, Sergey and Lu, Yao and Luu, Linda and
               Parada, Carolina and Pastor, Peter and Quiambao, Jornell and
               Rao, Kanishka and Rettinghouse, Jarek and Reyes, Diego and
               Sermanet, Pierre and Sievers, Nicolas and Tan, Clayton and
               Toshev, Alexander and Vanhoucke, Vincent and Xia, Fei and
               Xiao, Ted and Xu, Peng and Xu, Sichun and Yan, Mengyuan and
               Zeng, Andy},
  title     = {Do As {I} Can, Not As {I} Say: Grounding Language in Robotic
               Affordances},
  booktitle = {Conference on Robot Learning (CoRL)},
  series    = {Proceedings of Machine Learning Research},
  volume    = {205},
  pages     = {287--318},
  publisher = {PMLR},
  year      = {2022}
}

@inproceedings{kim2024openvla,
  author    = {Kim, Moo Jin and Pertsch, Karl and Karamcheti, Siddharth and
               Xiao, Ted and Balakrishna, Ashwin and Nair, Suraj and
               Rafailov, Rafael and Foster, Ethan and Lam, Grace and
               Sanketi, Pannag and Vuong, Quan and Kollar, Thomas and
               Burchfiel, Benjamin and Tedrake, Russ and Sadigh, Dorsa and
               Levine, Sergey and Liang, Percy and Finn, Chelsea},
  title     = {{OpenVLA}: An Open-Source Vision-Language-Action Model},
  booktitle = {Conference on Robot Learning (CoRL)},
  year      = {2024},
  note      = {arXiv:2406.09246}
}

@article{sutton1999options,
  author  = {Sutton, Richard S. and Precup, Doina and Singh, Satinder},
  title   = {Between {MDPs} and semi-{MDPs}: A framework for temporal
             abstraction in reinforcement learning},
  journal = {Artificial Intelligence},
  volume  = {112},
  number  = {1-2},
  pages   = {181--211},
  year    = {1999},
  doi     = {10.1016/S0004-3702(99)00052-1}
}

@inproceedings{lee2019transition,
  author    = {Lee, Youngwoon and Sun, Shao-Hua and Somasundaram, Sriram and
               Hu, Edward S. and Lim, Joseph J.},
  title     = {Composing Complex Skills by Learning Transition Policies},
  booktitle = {International Conference on Learning Representations (ICLR)},
  year      = {2019}
}

@inproceedings{byun2022distribution,
  author    = {Byun, Ju-Seung and Perrault, Andrew},
  title     = {Training Transition Policies via Distribution Matching for
               Complex Tasks},
  booktitle = {International Conference on Learning Representations (ICLR)},
  year      = {2022}
}

@article{xin2026seamless,
  author  = {Xin, Liming and Tian, Hanbin and Sheng, Bin},
  title   = {Seamless skill transitions with hierarchical reward shaping and
             failure-driven replay},
  journal = {Neurocomputing},
  volume  = {688},
  pages   = {133801},
  year    = {2026},
  doi     = {10.1016/j.neucom.2026.133801}
}

@inproceedings{lee2021tstar,
  author    = {Lee, Youngwoon and Lim, Joseph J. and Anandkumar, Anima and
               Zhu, Yuke},
  title     = {Adversarial Skill Chaining for Long-Horizon Robot Manipulation
               via Terminal State Regularization},
  booktitle = {Conference on Robot Learning (CoRL)},
  year      = {2021}
}

@inproceedings{chen2023seqdex,
  author    = {Chen, Yuanpei and Wang, Chen and Fei-Fei, Li and Liu, C. Karen},
  title     = {Sequential Dexterity: Chaining Dexterous Policies for
               Long-Horizon Manipulation},
  booktitle = {Conference on Robot Learning (CoRL)},
  year      = {2023}
}

@inproceedings{chen2024scar,
  author    = {Chen, Zixuan and Ji, Ze and Huo, Jing and Gao, Yang},
  title     = {{SCaR}: Refining Skill Chaining for Long-Horizon Robotic
               Manipulation via Dual Regularization},
  booktitle = {Advances in Neural Information Processing Systems (NeurIPS)},
  volume    = {37},
  year      = {2024}
}

@article{silver2018residual,
  author  = {Silver, Tom and Allen, Kelsey and Tenenbaum, Josh and
             Kaelbling, Leslie},
  title   = {Residual Policy Learning},
  journal = {arXiv preprint arXiv:1812.06298},
  year    = {2018}
}

@inproceedings{johannink2019residual,
  author    = {Johannink, Tobias and Bahl, Shikhar and Nair, Ashvin and
               Luo, Jianlan and Kumar, Avinash and Loskyll, Matthias and
               Ojea, Juan Aparicio and Solowjow, Eugen and Levine, Sergey},
  title     = {Residual Reinforcement Learning for Robot Control},
  booktitle = {IEEE International Conference on Robotics and Automation (ICRA)},
  pages     = {6023--6029},
  year      = {2019},
  doi       = {10.1109/ICRA.2019.8794127}
}

@article{tidd2022setup,
  author  = {Tidd, Brendan and Hudson, Nicolas and Cosgun, Akansel and
             Leitner, J{\"u}rgen},
  title   = {Learning Setup Policies: Reliable Transition Between Locomotion
             Behaviours},
  journal = {IEEE Robotics and Automation Letters},
  volume  = {7},
  number  = {4},
  pages   = {11958--11965},
  year    = {2022},
  doi     = {10.1109/LRA.2022.3207567}
}

@article{chai2025n2m,
  author  = {Chai, Kaixin and Lee, Hyunjun and Lim, Joseph J.},
  title   = {{N2M}: Bridging Navigation and Manipulation by Learning Pose
             Preference from Rollout},
  journal = {arXiv preprint arXiv:2509.18671},
  year    = {2025}
}

@article{yu2026skillswitch,
  author  = {Yu, Wanming and Acero, Fernando and Atanassov, Vassil and
             Yang, Chuanyu and Havoutis, Ioannis and Kanoulas, Dimitrios and
             Li, Zhibin},
  title   = {Discovery of skill-switching criteria for learning agile quadruped
             locomotion},
  journal = {Frontiers in Robotics and AI},
  volume  = {13},
  pages   = {1697159},
  year    = {2026},
  doi     = {10.3389/frobt.2026.1697159}
}

@article{kuang2025skillblender,
  author  = {Kuang, Yuxuan and Geng, Haoran and Elhafsi, Amine and
             Do, Tan-Dzung and Abbeel, Pieter and Malik, Jitendra and
             Pavone, Marco and Wang, Yue},
  title   = {{SkillBlender}: Towards Versatile Humanoid Whole-Body
             Loco-Manipulation via Skill Blending},
  journal = {arXiv preprint arXiv:2506.09366},
  year    = {2025}
}

@article{xin2026rpg,
  author  = {Xin, Yucheng and Bao, Jiacheng and Dong, Yubo and
             Wang, Xueqian and Zhao, Bin and Li, Xuelong and Tan, Junbo and
             Wang, Dong},
  title   = {{RPG}: Robust Policy Gating for Smooth Multi-Skill Transitions in
             Humanoid Fighting},
  journal = {arXiv preprint arXiv:2604.21355},
  year    = {2026}
}

@article{wu2026parkour,
  author  = {Wu, Zhen and Huang, Xiaoyu and Yang, Lujie and Zhang, Yuanhang and
             Chen, Xi and Abbeel, Pieter and Duan, Rocky and
             Kanazawa, Angjoo and Sferrazza, Carmelo and Shi, Guanya and
             Liu, C. Karen},
  title   = {Perceptive Humanoid Parkour: Chaining Dynamic Human Skills via
             Motion Matching},
  journal = {arXiv preprint arXiv:2602.15827},
  year    = {2026}
}

@inproceedings{zhang2021bisim,
  author    = {Zhang, Amy and McAllister, Rowan Thomas and Calandra, Roberto and
               Gal, Yarin and Levine, Sergey},
  title     = {Learning Invariant Representations for Reinforcement Learning
               without Reconstruction},
  booktitle = {International Conference on Learning Representations (ICLR)},
  year      = {2021}
}

@article{kumar2023cascaded,
  author  = {Kumar, K. Niranjan and Essa, Irfan and Ha, Sehoon},
  title   = {Cascaded Compositional Residual Learning for Complex Interactive
             Behaviors},
  journal = {IEEE Robotics and Automation Letters},
  volume  = {8},
  number  = {8},
  pages   = {4601--4608},
  year    = {2023},
  doi     = {10.1109/LRA.2023.3286171}
}

@inproceedings{mysore2022multicritic,
  author    = {Mysore, Siddharth and Cheng, George and Zhao, Yunqi and
               Saenko, Kate and Wu, Meng},
  title     = {Multi-Critic Actor Learning: Teaching {RL} Policies to Act with
               Style},
  booktitle = {International Conference on Learning Representations (ICLR)},
  year      = {2022}
}

@article{harvey2020lafan1,
  author  = {Harvey, F{\'e}lix G. and Yurick, Mike and
             Nowrouzezahrai, Derek and Pal, Christopher},
  title   = {Robust Motion In-betweening},
  journal = {ACM Transactions on Graphics},
  volume  = {39},
  number  = {4},
  year    = {2020},
  doi     = {10.1145/3386569.3392480}
}

@article{hallak2015contextual,
  title={Contextual markov decision processes},
  author={Hallak, Assaf and Di Castro, Dotan and Mannor, Shie},
  journal={arXiv preprint arXiv:1502.02259},
  year={2015}
}

\end{document}